\documentclass{article}

\usepackage[eandd, final]{neurips_2026}

\usepackage[utf8]{inputenc}
\usepackage[T1]{fontenc}
\usepackage[hyperfootnotes=false]{hyperref}
\usepackage{booktabs}
\usepackage{microtype}
\usepackage[dvipsnames]{xcolor}
\usepackage{multirow}
\usepackage{graphicx}
\usepackage{subcaption}
\usepackage{adjustbox}
\usepackage{cleveref}

\renewcommand{\arraystretch}{1.12}
\newcommand{\algoname}{OverLay++}
\makeatletter
\newcommand{\customfootnotetext}[2]{%
  \begingroup
  \setlength{\skip\footins}{0pt}
  \renewcommand{\@makefntext}[1]{%
    \microtypesetup{protrusion=false}%
    \setlength{\parindent}{0pt}%
    \setlength{\leftskip}{1.8em}%
    \noindent\llap{\@makefnmark}##1%
  }%
  \renewcommand{\thefootnote}{#1}%
  \footnotetext{#2}%
  \endgroup
}
\makeatother
\usepackage{tcolorbox}
\tcbuselibrary{breakable, skins}

\tcbset{
    promptbox/.style={
        breakable,
        enhanced,
        colback=gray!8,
        colframe=gray!40,
        fonttitle=\small\bfseries,
        fontupper=\small\ttfamily\raggedright,
        coltitle=black,
        boxrule=0.4pt,
        arc=3pt,
        left=6pt, right=6pt, top=4pt, bottom=4pt,
    }
}

\title{OverLay++: Dense-Overlap Layout-to-Image Generation Dataset}

\author{%
    \normalfont Shivansh Aggarwal\textsuperscript{1,\ensuremath{*}} \quad
    Shresth Grover\textsuperscript{1,\ensuremath{*}} \quad
    Divyansh Srivastava\textsuperscript{1,\ensuremath{*}} \\
    Haiyang Xu\textsuperscript{1} \quad Bingnan Li\textsuperscript{1} \quad
    Xiang Zhang\textsuperscript{1} \quad Ethan J. Armand\textsuperscript{1} \\
    Chuan Li\textsuperscript{2} \quad Jianwen Xie\textsuperscript{2} \quad
    Zhuowen Tu\textsuperscript{1} \\[4pt]
    \textsuperscript{1}UC San Diego \quad \textsuperscript{2}Lambda, Inc. \quad
    \textsuperscript{\ensuremath{*}}Equal contribution
}

\hypersetup{
    hidelinks,
    pdftitle={OverLay++: Dense-Overlap Layout-to-Image Generation Dataset},
    pdfauthor={Shivansh Aggarwal, Shresth Grover, Divyansh Srivastava, Haiyang Xu, Bingnan Li, Xiang Zhang, Ethan J. Armand, Chuan Li, Jianwen Xie, Zhuowen Tu}
}

\begin{document}
\raggedbottom

\maketitle

\customfootnotetext{}{%
    \makebox[4em][l]{Website:}{\scriptsize \url{https://mlpc-ucsd.github.io/OverLayPP}}\newline
    \makebox[4em][l]{Dataset:}{\scriptsize \url{https://huggingface.co/datasets/mlpcucsd/OverLayPP}}
}

\begin{figure}[h]
    \centering
    \includegraphics[width=\linewidth]{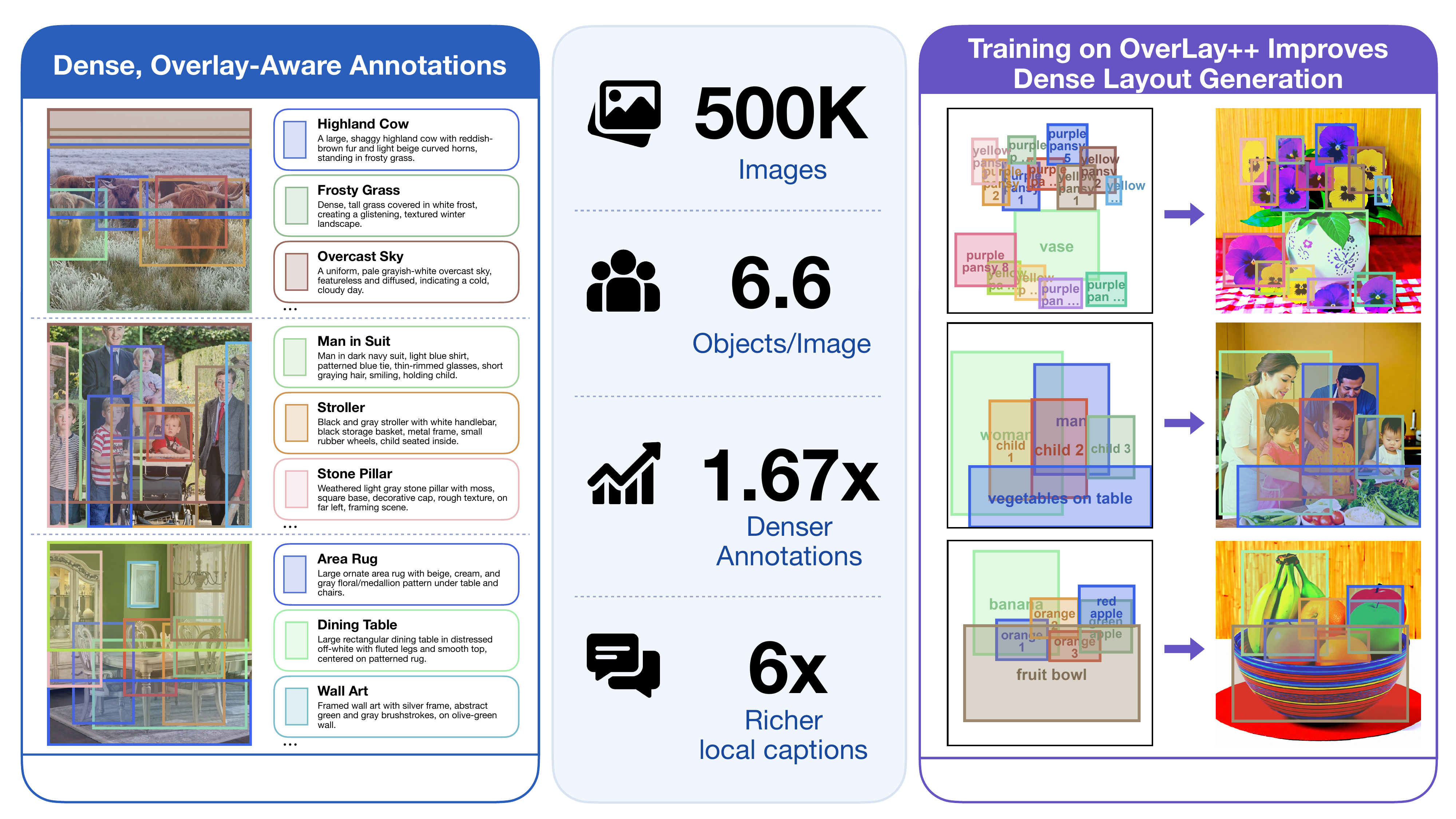}
    \caption{\textbf{OverLay++: A dense and overlap-aware dataset for Layout-to-Image generation.} Left: OverLay++ provides dense object annotations with rich local captions for fine-grained spatial and attribute control. Center: Our dataset contains 500K images with 6.6 objects per image, yielding denser annotations and richer captions. Right: Training on OverLay++ improves generation quality in dense and overlapping scenes, producing better object fidelity and spatial alignment.}
    \label{fig:teaser}
\end{figure}

\begin{abstract}
    Layout-to-Image generation has made substantial progress in spatial and object-level control. However, existing methods still struggle with complex scenes containing many overlapping and interacting objects. We argue that training data is a particular bottleneck: existing datasets lack examples with dense, complex object interactions. To address this gap, we introduce \textbf{\algoname{}}, a large-scale Layout-to-Image dataset with structurally complex scenes. \algoname{} contains approximately 500K images with an average of 6.6 objects per image, exceeding existing datasets by $1.67\times$ in annotation density. Beyond annotation density, \algoname{} provides rich semantic detail with object captions over six times longer than in current datasets. Our dataset generation pipeline is simple and produces dense, overlapping object annotations with rich per-object captions. Across multiple benchmarks, state-of-the-art Layout-to-Image methods trained on the \algoname{} dataset show consistent improvement and faster convergence, demonstrating the importance of dense, overlap-aware, and caption-rich supervision for controllable image generation.
\end{abstract}

\section{Introduction}
\label{sec:intro}

While text-to-image diffusion models~\citep{imagen,ramesh2021zero,nichol2021glide,rombach2022high,chen2023pixart,xue2024raphael} have achieved remarkable progress in generating high-quality and realistic images, they provide imperfect control over visual attributes such as spatial arrangement, color, and texture. This has led to a growing interest in controllable image generation~\citep{li2023gligenopensetgroundedtexttoimage,zhang2023adding,xie2023boxdiff,wang2024instancediffusion,lee2024groundit,chen2025ragd,chen2024unirealuniversalimagegeneration,tan2025ominicontrolminimaluniversalcontrol,zhao2025diceptiongeneralistdiffusionmodel,binyamin2024make,yang2023reco,srivastava2025dpardynamicpatchificationefficient}, where users provide additional conditions, such as depth maps~\citep{zhou20243dis,zhou20253disflux}, human pose~\citep{zhang2023adding}, or scene layouts~\citep{li2023gligenopensetgroundedtexttoimage,zhang2025creatilayoutsiamesemultimodaldiffusion}, to guide generation. Among these conditions, scene layouts provide a direct way to specify object locations, counts, and object-level attributes, thereby supporting a broad range of controllable image generation scenarios. Specifically, in Layout-to-Image generation~\citep{zhang2025creatilayoutsiamesemultimodaldiffusion,zhang2025eligenentitylevelcontrolledimage,wang2024instancediffusion,li2023gligenopensetgroundedtexttoimage,zhang2023adding,zhou2024migcmultiinstancegenerationcontroller,lee2024groundit}, the model is conditioned on a global text caption and a scene layout. Each layout element consists of a bounding box that specifies the object's location and a local text caption that describes the object.

Layout-to-Image generation methods typically adapt diffusion models~\citep{scorebased,ddpm,ddim,adm,cfg,rombach2022high,dalle2,imagen,esser2024scaling,srivastava2025layyourscenenaturalscenelayout} pretrained on internet-scale datasets~\citep{schuhmann2022laion5bopenlargescaledataset,gadre2023datacompsearchgenerationmultimodal,kakaobrain2022coyo-700m} to support layout conditioning. Recent methods~\citep{zhang2025creatilayoutsiamesemultimodaldiffusion,zhang2025eligenentitylevelcontrolledimage} use stronger backbones such as SD3~\citep{esser2024scaling} and FLUX~\citep{BlackForestLabs2024Flux}, where multimodal DiT blocks~\citep{peebles2023scalable} allow layout, visual, and text information to be combined within a shared architecture. However, current methods still struggle in complex scenes with many objects, particularly when objects overlap~\citep{li2025overlaybenchbenchmarklayouttoimagegeneration}. We argue that these failures reflect limitations in existing training data: current datasets contain few dense, overlapping layouts, and their local captions often omit object-level details needed to control color, texture, and interactions.

Existing datasets for Layout-to-Image generation are largely adapted from object detection~\citep{lin2015microsoftcococommonobjects,shao2019objects365} and visual grounding datasets~\citep{plummer2015flickr30k,Kosmos2}, which contain relatively few dense object interactions. Further, object annotations typically use class names or short phrases, providing little detail about object attributes. More recent data-collection efforts, including the dataset proposed by EliGen~\citep{zhang2025eligenentitylevelcontrolledimage} and LayoutSAM~\citep{zhang2025creatilayoutsiamesemultimodaldiffusion}, improve these annotations and object density with foundation models such as GroundingDINO~\citep{liu2023grounding} for improved object grounding and Qwen2-VL~\citep{wang2024qwen2} for detailed object annotations. However, these data-generation pipelines do not specifically target scenes with dense, overlapping objects, leaving such scenes underrepresented in the resulting datasets.

To address this gap, we introduce \textbf{\algoname{}}, a large-scale dataset for dense and overlap-aware Layout-to-Image generation. \algoname{} contains approximately 500K images with an average of 6.6 objects per image, along with short and long captions at both the global and per-object levels. Notably, even our short captions are $6\times$ longer than those in existing datasets. We construct \algoname{} using a simple and robust pipeline that prompts MLLMs to generate overlapping object proposals and applies per-object grounded validation. This produces images with dense layouts, substantial overlap, rich captions, and high instance-level accuracy. Training state-of-the-art L2I methods on \algoname{} improves performance across multiple benchmarks, and we find that models trained on \algoname{} converge faster than corresponding baselines trained on existing datasets, highlighting the value of dense, overlap-aware, and caption-rich supervision.

Our contributions are summarized as follows:
\begin{enumerate}
    \item We introduce \textbf{\algoname{}}, a large-scale dataset for dense and overlap-aware Layout-to-Image generation. \algoname{} contains approximately 500K images with an average of 6.6 objects per image, along with short and long captions at both the global and per-object levels. Notably, the short captions in \algoname{} are $6\times$ longer than those in existing datasets.

    \item We design a simple and robust dataset generation pipeline that targets dense, complex scenes. The pipeline prompts MLLMs to generate overlapping object proposals and applies per-object grounded validation to improve accuracy and localization.

    \item We benchmark state-of-the-art Layout-to-Image generation methods trained on \algoname{} and show improved performance across multiple benchmarks. We further find that models trained on \algoname{} converge faster than the corresponding baselines trained on existing datasets. Our results highlight the importance of dense layouts, substantial object overlap, and rich captions.
\end{enumerate}

\section{Related Work}

\subsection{Layout-to-Image Generation}

Layout-to-Image (L2I) makes Text-to-Image (T2I) models more controllable for real-world applications, enabling image synthesis that is both visually realistic and spatially grounded. L2I methods typically adapt text-to-image diffusion models~\citep{scorebased,ddpm,ddim,adm,cfg,rombach2022high,dalle2,imagen,esser2024scaling,srivastava2025layyourscenenaturalscenelayout}, pretrained on internet-scale datasets~\citep{schuhmann2022laion5bopenlargescaledataset,gadre2023datacompsearchgenerationmultimodal,kakaobrain2022coyo-700m}. Broadly, existing L2I methods can be divided into training-free and training-based approaches. Training-free methods~\citep{lee2024groundit,chen2025ragd,bar2023multidiffusion,kim2023dense,chefer2023attend,xie2023boxdiff,llmblueprint,rpg,trainigfreeregional} avoid modifying the underlying T2I model and instead encourage objects to appear within their intended spatial regions through inference modification (e.g., altering attention maps or modifying classifier-free guidance). While attractive for their simplicity and compatibility with off-the-shelf models, these methods provide limited control in complex scenes and perform worse than training-based counterparts~\citep{li2025overlaybenchbenchmarklayouttoimagegeneration}.

On the other hand, training-based methods modify the underlying T2I architecture to incorporate layout conditions. Early L2I methods~\citep{li2023gligenopensetgroundedtexttoimage,wang2024instancediffusion,zhou2024migcmultiinstancegenerationcontroller,cheng2024hicohierarchicalcontrollablediffusion} were largely built on latent-diffusion U-Net~\citep{ronneberger2015u} backbones and introduced layout information through layout-specific adapters, spatial masks, gated attention layers, or ControlNet~\citep{zhang2023adding}-like module additions. For example, GLIGEN~\citep{li2023gligenopensetgroundedtexttoimage} adds gated self-attention layers that incorporate bounding-box conditions into a pretrained SD-1.5 diffusion model. Similarly, InstanceDiffusion~\citep{wang2024instancediffusion} incorporates various spatial conditioning schemes with specialized fusion layers. Although these methods demonstrated the effectiveness of layout conditioning, their generation quality and controllability are constrained by the representational capacity of U-Net backbones, especially in dense scenes. Recent L2I methods leverage stronger multimodal DiT~\citep{peebles2023scalable} backbones, such as SD3~\citep{esser2024scaling} and FLUX~\citep{BlackForestLabs2024Flux}. Compared with earlier U-Net-based models, these backbones offer greater modeling capacity and improved image realism. As a result, recent works incorporate layout information into DiTs through text-like layout tokens, region-level attention masks, or multimodal conditioning mechanisms~\citep{zhang2025creatilayoutsiamesemultimodaldiffusion,zhang2025eligenentitylevelcontrolledimage}. While there has been significant progress in algorithm design, these models still struggle to handle complex scenes with numerous overlapping objects. Our work instead focuses on improving the training data by constructing denser layouts with higher levels of object overlap and richer local and global descriptions, ensuring that L2I models can learn complex object interactions and generate visually and spatially coherent images.

\subsection{Training Datasets for Layout-to-Image Generation}

Early Layout-to-Image (L2I) generation methods, including GLIGEN~\citep{li2023gligenopensetgroundedtexttoimage}, were typically trained on object detection~\citep{shao2019objects365,lin2015microsoftcococommonobjects} and grounding datasets. While these datasets provide bounding-box annotations, their localized annotations are usually class labels rather than descriptive object-level captions. This sparsity limits their utility for training L2I models, where fine-grained object descriptions are important for controllable generation.

To address this limitation, InstanceDiffusion~\citep{wang2024instancediffusion} generated dense object-level captions by first extracting objects with pretrained tagging models~\citep{zhang2024recognize}, detecting objects with Grounding-SAM~\citep{ren2024grounded}, and then captioning each detected object using BLIP-2~\citep{li2022blip,li2023blip}. More recent works, including SiamLayout~\citep{zhang2025creatilayoutsiamesemultimodaldiffusion} and EliGen~\citep{zhang2025eligenentitylevelcontrolledimage}, further improve L2I datasets by refining data construction pipelines and leveraging stronger foundation models. SiamLayout derived base images from the SA-1B~\citep{kirillov2023segany} dataset, using GroundingDINO~\citep{liu2023grounding} for object localization and VLMs~\citep{yao2024minicpm} for generating global and local captions. In contrast, EliGen relies on synthetic images: it samples captions from DiffusionDB~\citep{wang2023diffusiondb}, generates images with FLUX.1-dev~\citep{BlackForestLabs2024Flux}, and then applies Qwen2-VL~\citep{wang2024qwen2} to re-caption each image, extract spatial locations of objects, and generate a caption for each object. While these methods substantially improve caption quality and regional descriptions, the resulting annotations still have limited structural complexity, particularly in dense scenes with many overlapping objects. Our dataset generation addresses this limitation by explicitly prompting VLMs to generate object proposals with overlapping bounding boxes, ensuring that the resulting images contain dense scenes with many objects, substantial overlap, and rich captions. Further, our dataset generation pipeline includes per-object grounded validation, ensuring high instance-level accuracy in the images.
\section{Dataset Curation Pipeline}
\label{sec:methodology}

\begin{figure}[t]
    \centering
    \includegraphics[width=\linewidth]{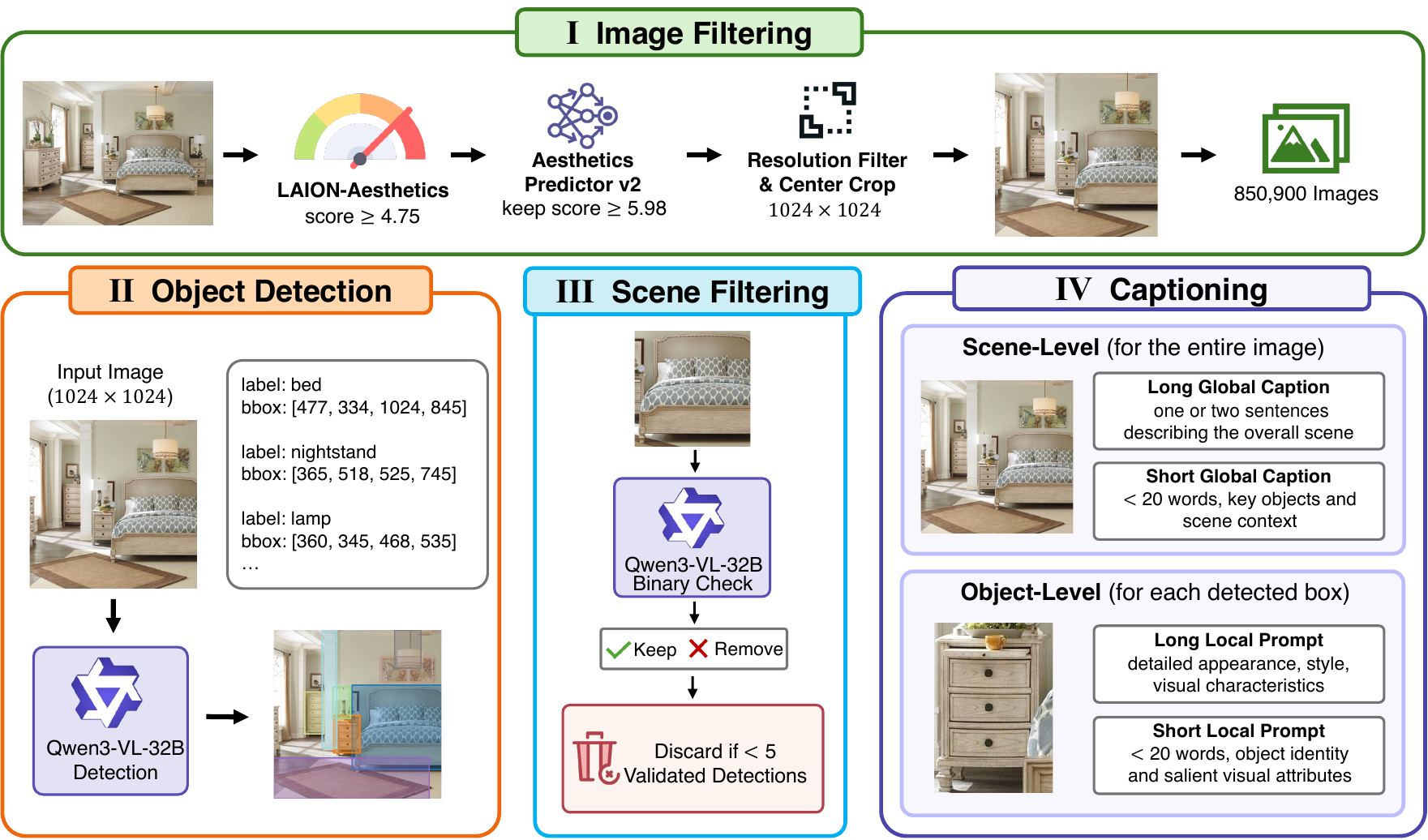}
    \caption{
        \textbf{Dataset generation pipeline for the \algoname{} dataset.}
        Four stages: (I) image filtering via aesthetics, resolution, and center cropping; (II) object detection for bounding boxes and labels; (III) scene filtering to remove low-quality or inconsistent samples, and (IV) hierarchical captioning producing scene-level and object-level captions (long and short).
    }
\end{figure}

\subsection{Motivation and Design Goals}

Existing layout-to-image datasets such as LayoutSAM \citep{zhang2025creatilayoutsiamesemultimodaldiffusion}, EliGen \citep{zhang2025eligenentitylevelcontrolledimage}, and GRIT \citep{Kosmos2} contain limited examples of complex scenes.

First, many commonly used datasets contain relatively sparse object layouts. As shown in Table~\ref{tab:dataset_analysis}, prior datasets include only 1.78--3.93 objects per image on average, which limits their ability to supervise dense multi-object generation. Beyond object number, we directly measure object density: for each bounding box, we identify its nearest neighbor based on center-to-center distance, then measure the intersection-over-union (IoU) for that pair. Mean NN IoU reflects the degree of overlap between nearby objects, while Mean NN Dist measures normalized spatial proximity in $[0,1]$ bounding box space. Existing datasets generally exhibit low NN IoU and relatively large NN distance, indicating that nearby objects often have weak local spatial interactions.

Second, existing datasets usually provide limited language supervision. As summarized in Table~\ref{tab:dataset_analysis}, most prior datasets rely primarily on image-level captions or sparse textual descriptions, without systematically providing object-level appearance descriptions tied to individual bounding boxes. This weakens supervision for fine-grained object attributes and makes it difficult for models to jointly learn global scene semantics and local object appearance from layout-conditioned data.

Based on these observations, our design goals are to construct a Layout-to-Image dataset with denser object layouts, stronger local spatial interactions, and richer multi-level language supervision. As shown in Table~\ref{tab:dataset_analysis}, our dataset substantially improves these key properties over existing datasets, including object density, nearest-neighbor spatial interaction, and annotation granularity. The detailed curation and annotation procedures used to achieve these improvements are described in the following subsections. Specifically, we develop an automated pipeline based on the LAION-Aesthetics V2~\citep{schuhmann2022laion5bopenlargescaledataset} 4.75 subset (956M images), consisting of four stages: 1) Image Filtering, 2) Object Detection, 3) Scene Filtering and 4) Captioning.

\begin{table}[th]
    \centering
    \small
    \setlength{\tabcolsep}{4pt}
    \renewcommand{\arraystretch}{1.25}
    \caption{
        \textbf{Dataset statistics for Layout-to-Image datasets.}
        Our dataset has higher object density, stronger spatial interactions, and richer caption supervision than prior datasets. Caption lengths are measured in number of words. \textbf{Bold} and \underline{underline} mark the best and second-best values in each column.
    }
    \label{tab:dataset_analysis}
    \begin{adjustbox}{max width=\linewidth}
        \begin{tabular}{@{}lrrrrrrrr@{}}
            \toprule
             &
             &
             &
             &
             &
            \multicolumn{2}{c}{\textbf{Local Caption}}
             &
            \multicolumn{2}{c}{\textbf{Global Caption}} \\
            \cmidrule(lr){6-7} \cmidrule(lr){8-9}
            \textbf{Dataset}
             & \textbf{Images}
             & \textbf{Objects} $\uparrow$
             & \textbf{IoU} $\uparrow$
             & \textbf{Dist} $\downarrow$
             & \textbf{Short} $\uparrow$
             & \textbf{Long} $\uparrow$
             & \textbf{Short} $\uparrow$
             & \textbf{Long} $\uparrow$                 \\
            \midrule

            LayoutSAM~\citep{zhang2025creatilayoutsiamesemultimodaldiffusion}
             & \underline{2.66M}
             & \underline{3.93}
             & 0.22
             & \underline{0.22}
             & 1.99
             & \underline{11.93}
             & --
             & \textbf{81.81}                           \\

            EliGen~\citep{zhang2025eligenentitylevelcontrolledimage}
             & 496K
             & 2.56
             & \underline{0.26}
             & 0.24
             & 2.12
             & --
             & \underline{17.76}
             & --                                       \\

            GRIT~\citep{Kosmos2}
             & \textbf{20.5M}
             & 1.78
             & 0.16
             & 0.25
             & \underline{2.40}
             & --
             & 17.47
             & --                                       \\

            \midrule
            \textbf{\algoname{} (Ours)}
             & 500K
             & \textbf{6.59}
             & \textbf{0.31}
             & \textbf{0.17}
             & \textbf{15.89}
             & \textbf{89.70}
             & \textbf{19.98}
             & \underline{52.24}                        \\

            \bottomrule
        \end{tabular}
    \end{adjustbox}
\end{table}

\subsection{Image Filtering}

Given the highly variable quality of web-scraped data, we first filter the corpus using the LAION-Aesthetics Predictor V2~\citep{schuhmann2022laion5bopenlargescaledataset}, retaining only images with an aesthetic score of at least $5.98$. This step ensures that the retained images have strong compositional and perceptual quality, providing a high-quality visual basis for dense layout construction. We additionally discard any image where either spatial dimension falls below 1024 pixels, enforcing $\min(H, W) \geq 1024$. Images that satisfy both criteria are then center-cropped deterministically to a standardized resolution of $1024 \times 1024$ pixels. After this filtering step, we obtain 850,900 images.

\subsection{Object Detection}

To address the lack of dense objects in existing datasets, we explicitly construct object-dense scenes. We prompt the model, Qwen3-VL-32B~\citep{bai2025qwen3vl}, to detect all visually salient objects and return structured predictions consisting of: (i) a semantic category label, (ii) a bounding box in $[x_1, y_1, x_2, y_2]$ format normalized to a 1024-pixel coordinate space, and (iii) a detailed caption describing each object's visual attributes. After the detection result, to further encourage dense layouts, images with fewer than five valid detections after applying these constraints are discarded.

\subsection{Scene Filtering}

To further reduce hallucinated or low-quality detections, we perform per-object validation. For each candidate detection, the corresponding image region is cropped and re-evaluated using Qwen3-VL-32B with a binary prompt that asks whether the described object is present in the cropped region. Detections for which the model returns a negative response are removed. Afterwards, to ensure the density of the objects, images with fewer than five remaining valid detections after this validation step are discarded. 499,249 images are kept after this step.

\subsection{Captioning}

We use Qwen3-VL-32B to do scene-level captioning. The retained images are annotated with complementary textual fields at two levels of granularity: scene-level descriptions and object-level descriptions. Scene-level captions provide holistic semantic context for the full image, while object-level captions describe the appearance and attributes of individual objects associated with specific bounding boxes.

While detailed captions are a valuable supervision signal, some layout-to-image methods use language encoders with short token limits~\citep{dalle2}. To increase dataset accessibility while preserving semantic supervision, we provide both \textit{long} and \textit{short} captions.

A \textit{long scene-level caption} is prompted to describe the overall image, including its main objects and their attributes, whereas a \textit{short scene-level caption} is prompted to use fewer than 20 words and provides a compact summary of the key objects and context. Similarly, the \textit{long object-level caption}, generated during the object detection stage, provides a detailed description of each object's appearance, style, and visual characteristics. In contrast, the corresponding \textit{short object-level caption}, also prompted to use fewer than 20 words, captures the object's identity and salient visual attributes. Together, these annotations provide both scene-level and object-level supervision. This design directly addresses the limitations of prior datasets by producing layouts with more objects, stronger local spatial interactions, and richer hierarchical annotations. The prompts are provided in Appendix~A.1. The word limits are prompt instructions rather than post-generation truncation rules, and Table~\ref{tab:dataset_analysis} reports observed caption lengths.

\section{Evaluation}
\label{sec:experiments}

\subsection{Implementation Details}

We evaluate the \algoname{} dataset using two state-of-the-art Layout-to-Image generation methods, EliGen~\citep{zhang2025eligenentitylevelcontrolledimage} and SiamLayout~\citep{zhang2025creatilayoutsiamesemultimodaldiffusion}. For EliGen, we use the publicly released code to train separate layout-conditioned models on the EliGen and \algoname{} datasets, starting from a pretrained SD3 backbone. Both models are trained for 10K steps with a per-GPU batch size of 12 on 8$\times$A6000 GPUs and a learning rate of $3 \times 10^{-4}$. The \algoname{} EliGen-SD3 run uses short local captions and long global captions. For SiamLayout, since the training code is not publicly available, we fine-tune the released model on \algoname{} for 60K steps with a per-GPU batch size of 2, gradient accumulation of 2, and a learning rate of $5 \times 10^{-5}$ on 8$\times$A6000 GPUs. We compare against the SiamLayout-SD3 results reported in the original paper~\citep{zhang2025creatilayoutsiamesemultimodaldiffusion}. All models are optimized with the AdamW~\citep{loshchilov2019decoupledweightdecayregularization} optimizer with a weight decay of 0.05, $\beta_1=0.9$, and $\beta_2=0.95$. We use a constant learning rate and a classifier-free guidance~\citep{cfg} scale of 7.5 for generation. We compare our dataset on three benchmarks: LayoutSAM~\citep{zhang2025creatilayoutsiamesemultimodaldiffusion}, OverLayBench~\citep{li2025overlaybenchbenchmarklayouttoimagegeneration}, and DenseLayout~\citep{xiang2025instanceassemble}.

\begin{figure*}[t]
    \centering
    \includegraphics[width=\linewidth]{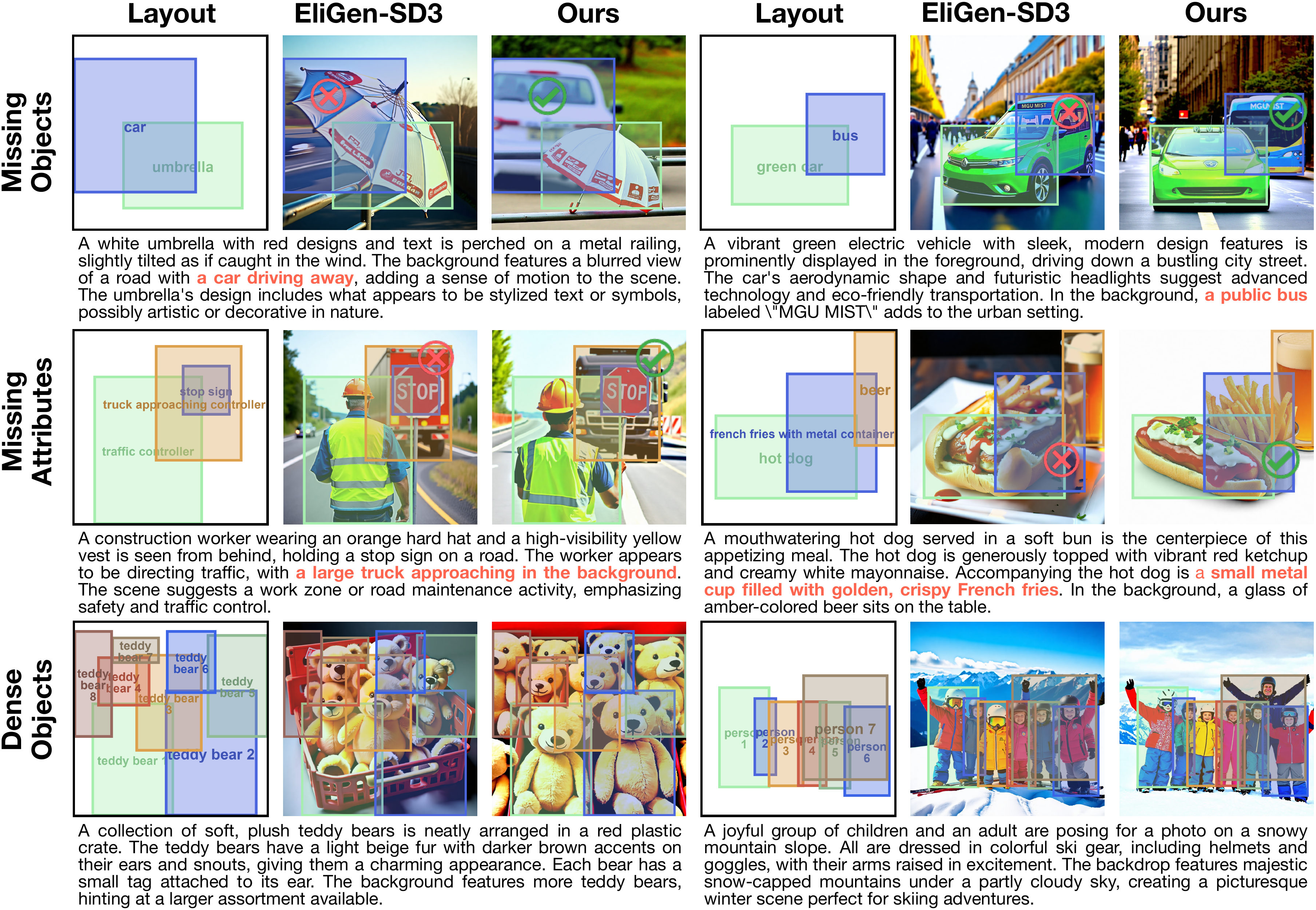}
    \caption{
        \textbf{Qualitative comparison of L2I generation.}
        Models trained on \algoname{} produce more coherent and detailed images than those trained on prior datasets. Common failure cases of prior datasets include missing objects, missing attributes, and poor handling of dense object layouts.
    }
    \label{fig:dataset_comparison}
\end{figure*}

\subsection{LayoutSAM Benchmark}
\begin{table}[t]
    \centering
    \small
    \setlength{\tabcolsep}{3pt}
    \renewcommand{\arraystretch}{1.25}
    \caption{
        \textbf{Results on LayoutSAM-Eval}~\citep{zhang2025creatilayoutsiamesemultimodaldiffusion}.
        Comparison of region-wise and global quality across different methods. Our methods consistently outperform their corresponding baselines, achieving substantial gains in region-wise quality while maintaining competitive global image quality. Improvements over baseline are shown in {\color{ForestGreen}green}, and \textbf{bold} and \underline{underline} denote the best and second-best results among all compared methods.
    }
    \label{tab:layoutsam_eval}
    \begin{adjustbox}{max width=\linewidth}
        \begin{tabular}{lcccccc}
            \toprule
             & \multicolumn{4}{c}{\textbf{Region-wise Quality}}
             & \multicolumn{2}{c}{\textbf{Global Quality}}                                                                                                       \\
            \cmidrule(lr){2-5} \cmidrule(lr){6-7}
            \textbf{Model}
             & \textbf{Spatial(\%)} $\uparrow$
             & \textbf{Color(\%)} $\uparrow$
             & \textbf{Texture(\%)} $\uparrow$
             & \textbf{Shape(\%)} $\uparrow$
             & \textbf{FID} $\downarrow$
             & \textbf{IS} $\uparrow$                                                                                                                            \\
            \midrule
            GLIGEN~\citep{li2023gligenopensetgroundedtexttoimage}
             & 77.53                                            & 49.41             & 55.29             & 52.72             & 21.92             & 20.57          \\
            Ranni~\citep{feng2023ranni}
             & 41.38                                            & 24.10             & 25.57             & 23.35             & 27.24             & 19.81          \\
            MIGC~\citep{zhou2024migcmultiinstancegenerationcontroller}
             & 85.66                                            & 66.97             & 71.24             & 69.06             & 21.19             & 19.65          \\
            InstanceDiffusion~\citep{wang2024instancediffusion}
             & 87.99                                            & 69.16             & 72.78             & 71.08             & 19.67             & 20.02          \\
            Be Yourself~\citep{dahary2024yourselfboundedattentionmultisubject}
             & 53.99                                            & 31.73             & 35.26             & 32.75             & 28.10             & 17.98          \\
            HiCo~\citep{cheng2024hicohierarchicalcontrollablediffusion}
             & 87.04                                            & 69.19             & 72.36             & 71.10             & 22.61             & 20.15          \\
            \midrule
            SiamLayout-SD3~\citep{zhang2025creatilayoutsiamesemultimodaldiffusion}
             & 92.67                                            & 74.45             & 77.21             & 75.93             & \underline{19.10} & \textbf{22.04} \\
            SiamLayout-SD3 (Ours)
             & \underline{93.30} {\color{ForestGreen}(+0.68\%)}
             & 75.99 {\color{ForestGreen}(+2.07\%)}
             & 78.77 {\color{ForestGreen}(+2.02\%)}
             & 78.41 {\color{ForestGreen}(+3.27\%)}
             & \textbf{17.96} {\color{ForestGreen}(-5.97\%)}
             & \underline{21.60}                                                                                                                                 \\
            \midrule
            EliGen-SD3~\citep{zhang2025eligenentitylevelcontrolledimage}
             & 93.04                                            & \underline{82.96} & \underline{86.73} & \underline{86.35} & 25.17             & 19.21          \\
            EliGen-SD3 (Ours)
             & \textbf{95.45} {\color{ForestGreen}(+2.59\%)}
             & \textbf{90.03} {\color{ForestGreen}(+8.52\%)}
             & \textbf{93.14} {\color{ForestGreen}(+7.39\%)}
             & \textbf{92.81} {\color{ForestGreen}(+7.48\%)}
             & 19.61 {\color{ForestGreen}(-22.09\%)}
             & 19.54 {\color{ForestGreen}(+1.72\%)}                                                                                                              \\
            \bottomrule
        \end{tabular}
    \end{adjustbox}
\end{table}

LayoutSAM-Eval~\citep{zhang2025creatilayoutsiamesemultimodaldiffusion} is a Layout-to-Image generation benchmark built from a subset of LayoutSAM and contains 5,000 layouts. It evaluates generation quality from two perspectives: region-wise quality and global quality. Region-wise quality uses VLM-based VQA to measure whether each object appears in the correct bounding box and whether its attributes, such as color, texture, and shape, match the local descriptions. Global quality is summarized using FID~\citep{heusel2018ganstrainedtimescaleupdate} and Inception Score (IS)~\citep{salimans2016improvedtechniquestraininggans}, which measure distribution-level image fidelity and diversity.

Results are reported in \Cref{tab:layoutsam_eval}. Training on \algoname{} consistently improves each model over its corresponding baseline. {Eligen-SD3 (Ours)} achieves state-of-the-art results on all region-wise quality metrics, with gains of up to 8.5\%. {SiamLayout-SD3 (Ours)} achieves the best FID and the second-best IS among all methods, demonstrating the effectiveness of training with the \algoname{} dataset.

\subsection{OverLayBench Benchmark}
\begin{table*}[t]
    \centering
    \small
    \setlength{\tabcolsep}{4pt}
    \renewcommand{\arraystretch}{1.22}
    \caption{
        \textbf{Results on OverLayBench}~\citep{li2025overlaybenchbenchmarklayouttoimagegeneration}.
        Comparison of generation quality across different methods and complexity settings. Our methods consistently outperform their corresponding baselines, achieving substantial gains in region-aware metrics while maintaining competitive overall image fidelity.
        Improvements over the baseline are shown in {\color{ForestGreen}green}, and \textbf{bold} and \underline{underline} indicate the best and second-best results among all compared methods.
    }
    \label{tab:quantitative_results_all_training}
    \begin{adjustbox}{max width=\textwidth}
        \begin{tabular}{@{}lccccccc@{}}
            \toprule
            \textbf{Method}
             & \textbf{mIoU(\%)} $\uparrow$
             & \textbf{O-mIoU(\%)} $\uparrow$
             & \textbf{SR$_E$(\%)} $\uparrow$
             & \textbf{SR$_R$(\%)} $\uparrow$
             & \textbf{CLIP$_G$} $\uparrow$
             & \textbf{CLIP$_L$} $\uparrow$
             & \textbf{FID} $\downarrow$                                                                                                                                                \\
            \midrule

            \multicolumn{8}{c}{\textbf{OverLayBench-Simple}}                                                                                                                            \\
            \midrule
            GLIGEN~\citep{li2023gligenopensetgroundedtexttoimage}
             & 60.54                                            & 36.22             & 49.99             & 78.72             & 34.17             & 24.75             & 31.27             \\

            InstanceDiffusion~\citep{wang2024instancediffusion}
             & \textbf{71.21}                                   & \textbf{49.99}    & 77.71             & 87.49             & 34.25             & 27.69             & 36.17             \\

            MIGC~\citep{zhou2024migcmultiinstancegenerationcontroller}
             & 58.64                                            & 32.15             & 63.41             & 81.60             & 33.07             & 26.49             & 31.64             \\

            HiCo~\citep{cheng2024hicohierarchicalcontrollablediffusion}
             & 69.47                                            & \underline{47.23} & 67.75             & 86.08             & 35.25             & 27.04             & 29.21             \\

            3DIS~\citep{zhou20243dis}
             & 65.75                                            & 38.38             & \underline{86.24} & 86.98             & 35.85             & \textbf{29.67}    & 29.18             \\

            \midrule
            SiamLayout-SD3~\citep{zhang2025creatilayoutsiamesemultimodaldiffusion}
             & 58.78                                            & 32.52             & 72.34             & 84.45             & \textbf{37.29}    & 27.49             & \underline{27.51} \\

            SiamLayout-SD3 (Ours)
             & 61.01 {\color{ForestGreen}(+3.79\%)}
             & 37.23 {\color{ForestGreen}(+14.48\%)}
             & 75.19 {\color{ForestGreen}(+3.94\%)}
             & 85.60 {\color{ForestGreen}(+1.36\%)}
             & \underline{37.03}
             & 27.59 {\color{ForestGreen}(+0.36\%)}
             & \textbf{26.73} {\color{ForestGreen}(-2.84\%)}                                                                                                                            \\

            \midrule
            EliGen-SD3~\citep{zhang2025eligenentitylevelcontrolledimage}
             & 66.67                                            & 41.34             & 85.70             & \underline{88.67} & 37.01             & 28.57             & 30.29             \\

            EliGen-SD3 (Ours)
             & \underline{70.39} {\color{ForestGreen}(+5.58\%)}
             & 47.09 {\color{ForestGreen}(+13.91\%)}
             & \textbf{91.83} {\color{ForestGreen}(+7.15\%)}
             & \textbf{91.06} {\color{ForestGreen}(+2.70\%)}
             & 36.27
             & \underline{29.15} {\color{ForestGreen}(+2.03\%)}
             & 30.28 {\color{ForestGreen}(-0.03\%)}                                                                                                                                     \\

            \midrule\midrule

            \multicolumn{8}{c}{\textbf{OverLayBench-Regular}}                                                                                                                           \\
            \midrule
            GLIGEN~\citep{li2023gligenopensetgroundedtexttoimage}
             & 52.46                                            & 26.53             & 44.88             & 77.46             & 33.93             & 23.42             & 52.22             \\

            InstanceDiffusion~\citep{wang2024instancediffusion}
             & \underline{60.08}                                & \underline{34.15} & 72.51             & 83.36             & 33.09             & 26.19             & 59.73             \\

            MIGC~\citep{zhou2024migcmultiinstancegenerationcontroller}
             & 47.42                                            & 20.06             & 56.67             & 77.85             & 32.72             & 24.99             & 54.24             \\

            HiCo~\citep{cheng2024hicohierarchicalcontrollablediffusion}
             & 55.02                                            & 29.60             & 58.24             & 79.89             & 33.91             & 25.34             & 49.07             \\

            3DIS~\citep{zhou20243dis}
             & 55.66                                            & 27.29             & 80.80             & \underline{83.69} & 35.42             & \underline{28.12} & 48.56             \\

            \midrule
            SiamLayout-SD3~\citep{zhang2025creatilayoutsiamesemultimodaldiffusion}
             & 47.04                                            & 20.67             & 62.60             & 78.31             & \textbf{36.67}    & 25.55             & \underline{45.57} \\

            SiamLayout-SD3 (Ours)
             & 48.41 {\color{ForestGreen}(+2.91\%)}
             & 23.32 {\color{ForestGreen}(+12.82\%)}
             & 65.53 {\color{ForestGreen}(+4.68\%)}
             & 80.39 {\color{ForestGreen}(+2.66\%)}
             & 36.49
             & 25.66 {\color{ForestGreen}(+0.43\%)}
             & \textbf{45.37} {\color{ForestGreen}(-0.44\%)}                                                                                                                            \\

            \midrule
            EliGen-SD3~\citep{zhang2025eligenentitylevelcontrolledimage}
             & 56.79                                            & 29.24             & \underline{81.34} & 83.57             & \underline{36.54} & 27.20             & 49.15             \\

            EliGen-SD3 (Ours)
             & \textbf{62.40} {\color{ForestGreen}(+9.88\%)}
             & \textbf{36.66} {\color{ForestGreen}(+25.38\%)}
             & \textbf{89.45} {\color{ForestGreen}(+9.97\%)}
             & \textbf{86.75} {\color{ForestGreen}(+3.81\%)}
             & 35.64
             & \textbf{28.17} {\color{ForestGreen}(+3.57\%)}
             & 49.09 {\color{ForestGreen}(-0.12\%)}                                                                                                                                     \\

            \midrule\midrule

            \multicolumn{8}{c}{\textbf{OverLayBench-Complex}}                                                                                                                           \\
            \midrule
            GLIGEN~\citep{li2023gligenopensetgroundedtexttoimage}
             & 50.79                                            & 23.85             & 41.70             & 79.93             & 33.92             & 22.75             & 57.32             \\

            InstanceDiffusion~\citep{wang2024instancediffusion}
             & \underline{53.68}                                & \underline{25.63} & 66.02             & 80.34             & 32.33             & 25.53             & 66.32             \\

            MIGC~\citep{zhou2024migcmultiinstancegenerationcontroller}
             & 40.04                                            & 13.26             & 47.80             & 74.48             & 31.93             & 24.20             & 66.52             \\

            HiCo~\citep{cheng2024hicohierarchicalcontrollablediffusion}
             & 46.56                                            & 20.35             & 48.88             & 75.19             & 33.15             & 24.41             & 55.78             \\

            3DIS~\citep{zhou20243dis}
             & 50.65                                            & 21.75             & \underline{74.31} & 81.57             & 35.11             & \textbf{27.35}    & 54.90             \\

            \midrule
            SiamLayout-SD3~\citep{zhang2025creatilayoutsiamesemultimodaldiffusion}
             & 44.24                                            & 18.05             & 52.10             & 79.98             & \textbf{36.55}    & 24.76             & \underline{53.29} \\

            SiamLayout-SD3 (Ours)
             & 45.81 {\color{ForestGreen}(+3.55\%)}
             & 20.42 {\color{ForestGreen}(+13.13\%)}
             & 57.54 {\color{ForestGreen}(+10.44\%)}
             & 81.60 {\color{ForestGreen}(+2.03\%)}
             & 36.15
             & 24.97 {\color{ForestGreen}(+0.85\%)}
             & \textbf{52.21} {\color{ForestGreen}(-2.03\%)}                                                                                                                            \\

            \midrule
            EliGen-SD3~\citep{zhang2025eligenentitylevelcontrolledimage}
             & 50.53                                            & 22.97             & 73.75             & \underline{82.22} & \underline{36.49} & 26.32             & 53.58             \\

            EliGen-SD3 (Ours)
             & \textbf{57.90} {\color{ForestGreen}(+14.59\%)}
             & \textbf{30.99} {\color{ForestGreen}(+34.92\%)}
             & \textbf{83.30} {\color{ForestGreen}(+12.95\%)}
             & \textbf{85.59} {\color{ForestGreen}(+4.10\%)}
             & 35.46
             & \underline{26.97} {\color{ForestGreen}(+2.47\%)}
             & 53.63                                                                                                                                                                    \\

            \bottomrule
        \end{tabular}
    \end{adjustbox}
\end{table*}

OverLayBench~\citep{li2025overlaybenchbenchmarklayouttoimagegeneration} evaluates instance-level layout control under increasing scene complexity, stratified into {Simple}, {Regular}, and {Complex} splits. We use Qwen2.5-VL-32B~\citep{bai2025qwen25vltechnicalreport} to extract instance-level bounding boxes and compute mIoU and O-mIoU (overlap-specific IoU), and success rates {SR$_E$} and {SR$_R$} under strict and relaxed IoU thresholds, respectively. Scene-level quality is assessed via {CLIP$_G$}, instance-level semantic alignment via {CLIP$_L$}, and overall realism via FID~\citep{heusel2018ganstrainedtimescaleupdate}.

Results are reported in \Cref{tab:quantitative_results_all_training} across all three difficulty tiers. Training on \algoname{} yields consistent gains across both baselines, demonstrating the effectiveness of our dataset. Across all OverLayBench splits, {Eligen-SD3 (Ours)} improves spatial alignment over the original EliGen-SD3 and achieves the best results on several metrics. On {OverLayBench-Simple}, it obtains the best SR$_E$ (91.83\%) and SR$_R$ (91.06\%). On {OverLayBench-Regular}, it achieves the best mIoU (62.40), O-mIoU (36.66), and SR$_E$ (89.45\%). On {OverLayBench-Complex}, it again obtains the best mIoU (57.90), O-mIoU (30.99), and SR$_E$ (83.30\%). These results suggest that \algoname{} improves layout alignment, particularly in dense scenes with overlapping objects.

\begin{figure}[t]
    \centering
    \begin{subfigure}[c]{0.32\linewidth}
        \centering
        \includegraphics[width=\linewidth]{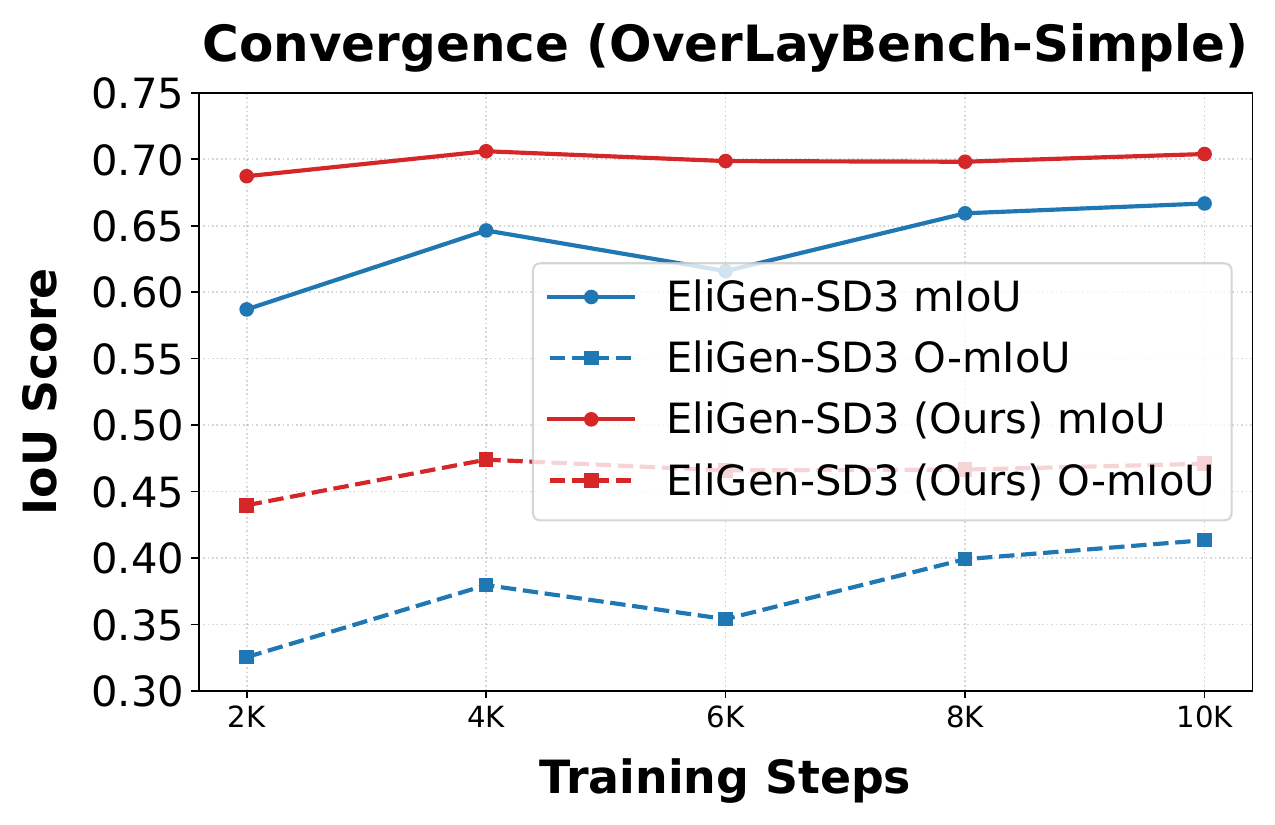}
        \caption{Simple}
    \end{subfigure}
    \hfill
    \begin{subfigure}[c]{0.32\linewidth}
        \centering
        \includegraphics[width=\linewidth]{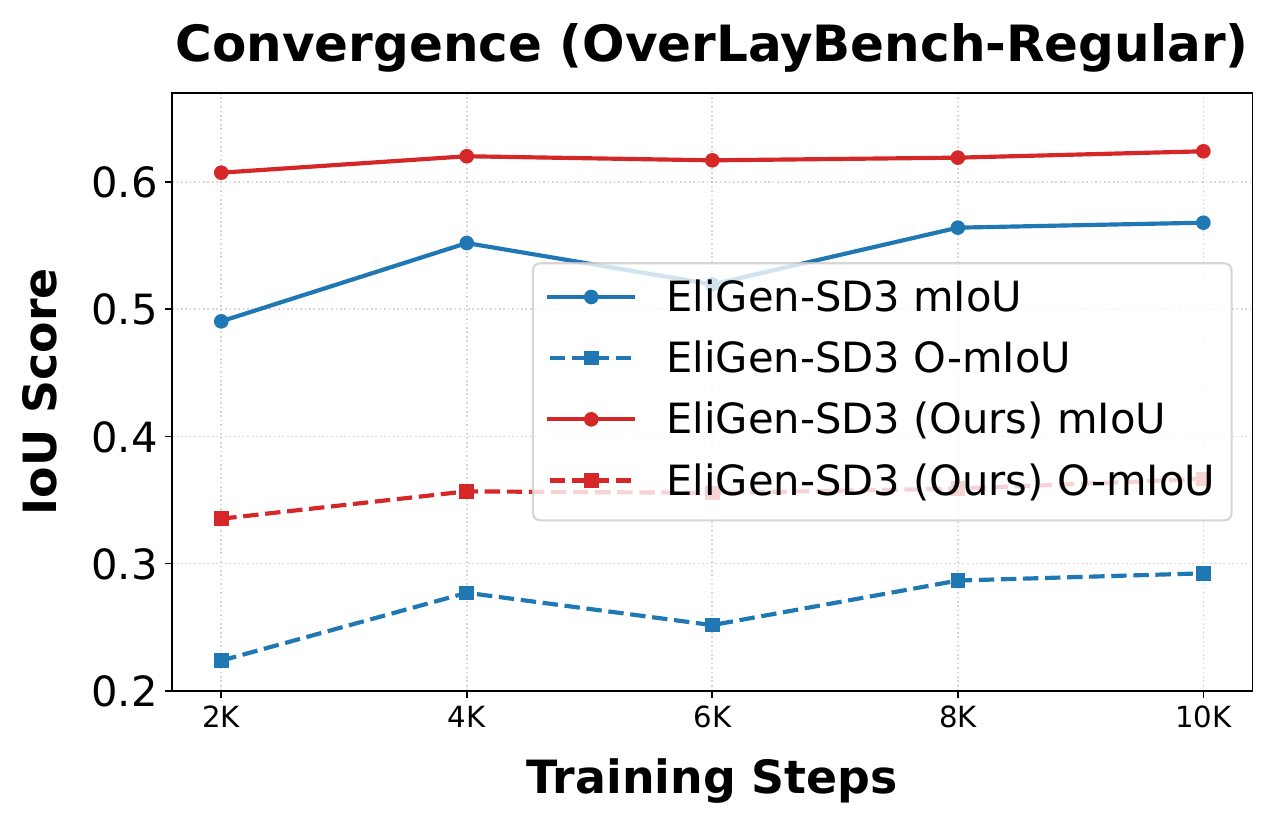}
        \caption{Regular}
    \end{subfigure}
    \hfill
    \begin{subfigure}[c]{0.32\linewidth}
        \centering
        \includegraphics[width=\linewidth]{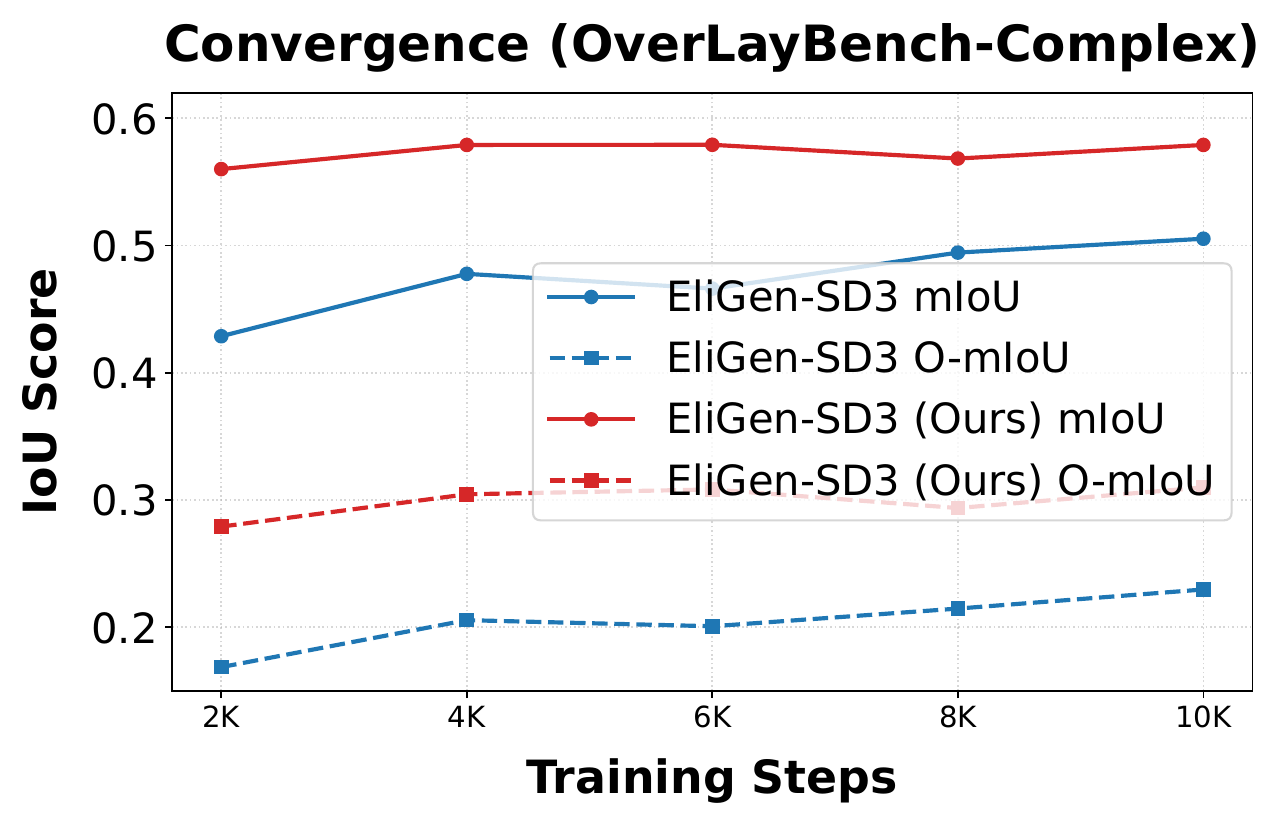}
        \caption{Complex}
    \end{subfigure}

    \caption{\textbf{Convergence analysis.}
        Comparison of mIoU and O-mIoU versus training steps for Simple, Regular, and Complex settings.
        The model trained on \algoname{} achieves higher scores across all difficulty levels and demonstrates stable convergence behavior.}
    \label{fig:convergence_all}
\end{figure}

We further compare the convergence of {EliGen-SD3} when trained on the EliGen dataset and on \algoname{}. As shown in \Cref{fig:convergence_all}, training on \algoname{} leads to faster IoU score convergence across all splits, suggesting that our dataset helps the model learn overlapping layouts more efficiently.

\subsection{DenseLayout Benchmark}
\begin{table}[t]
    \centering
    \caption{
        \textbf{Results on DenseLayout}~\citep{xiang2025instanceassemble}.
        EliGen-SD3 and SiamLayout-SD3 trained on \algoname{} outperform their corresponding baselines on DenseLayout benchmark.
        Improvements over baselines are highlighted in {\color{ForestGreen}green}, and \textbf{bold} indicates the best results among all compared methods.
    }
    \label{tab:instanceassemble_dense_layout}
    \setlength{\tabcolsep}{2.5pt}
    \renewcommand{\arraystretch}{1.15}
    \footnotesize
    \begin{tabular}{lcccc}
        \toprule
        \textbf{Model} & \textbf{mIoU(\%)} $\uparrow$                  & \textbf{Color(\%)} $\uparrow$ & \textbf{Texture(\%)} $\uparrow$ & \textbf{Shape(\%)} $\uparrow$ \\
        \midrule
        EliGen-SD3~\citep{zhang2025eligenentitylevelcontrolledimage}
                       & 45.79                                         & 37.65                         & 39.98                           & 39.76                         \\
        EliGen-SD3 (Ours)
                       & \textbf{46.92} {\color{ForestGreen}(+2.47\%)}
                       & \textbf{40.61} {\color{ForestGreen}(+7.86\%)}
                       & \textbf{42.91} {\color{ForestGreen}(+7.33\%)}
                       & \textbf{42.82} {\color{ForestGreen}(+7.70\%)}                                                                                                   \\

        \midrule
        SiamLayout-SD3~\citep{zhang2025creatilayoutsiamesemultimodaldiffusion}
                       & 15.54                                         & 11.69                         & 12.34                           & 12.17                         \\
        SiamLayout-SD3 (Ours)
                       & 17.10 {\color{ForestGreen}(+10.04\%)}
                       & 13.15 {\color{ForestGreen}(+12.49\%)}
                       & 13.77 {\color{ForestGreen}(+11.59\%)}
                       & 13.61 {\color{ForestGreen}(+11.83\%)}                                                                                                           \\
        \bottomrule
    \end{tabular}
\end{table}

DenseLayout~\citep{xiang2025instanceassemble} evaluates Layout-to-Image generation in high-density scenes with many objects and fine-grained spatial constraints. Following the Layout Grounding Score protocol used in LayoutSAM-Eval, GroundingDINO~\citep{liu2023grounding} detects each labeled instance and computes the {mIoU} score. Instance-level attribute alignment is then evaluated by a vision-language model~\citep{yao2024minicpm} on each cropped region, conditioned on the corresponding instance description, resulting in three attribute metrics: {Color}, {Texture}, and {Shape}.

Results are shown in \Cref{tab:instanceassemble_dense_layout}. Training on \algoname{} improves both EliGen-SD3 and SiamLayout-SD3 across all evaluation metrics. {EliGen-SD3 (Ours)} achieves the best overall performance, improving mIoU from 45.79 to 46.92 and consistently improving attribute alignment, with gains of 7.86\% in Color, 7.33\% in Texture, and 7.7\% in Shape. {SiamLayout-SD3 (Ours)} also improves over its baseline on all metrics, showing that \algoname{} benefits dense-layout generation across different L2I models.

\subsection{Caption Granularity}

We ablate the effect of caption granularity by training {EliGen-SD3} with different combinations of short and long captions at the local and global levels. Results are shown in \Cref{tab:ablation_prompt_granularity}. The {Short-Long} setting achieves the best overall results, suggesting that short local captions are more effective for object grounding while long global captions provide useful scene-level context. This is likely due to the CLIP text encoder used for local prompts, which is limited to 77 tokens and may truncate longer object descriptions. With stronger local text encoders or longer context windows, long local captions may provide additional gains.

\begin{table*}[t]
    \centering
    \small
    \setlength{\tabcolsep}{4pt}
    \renewcommand{\arraystretch}{1.15}
    \caption{
        \textbf{Ablation study on caption granularity on OverLayBench.} Short local captions with long global captions achieve the best region-level alignment. \textbf{Bold} and \underline{underline} indicate the best and second-best results within each split.
    }
    \label{tab:ablation_prompt_granularity}
    \begin{adjustbox}{max width=\textwidth}
        \begin{tabular}{@{}llllcccccc@{}}
            \toprule
            \textbf{Method}
             & \textbf{Split}
             & \textbf{Local Caption}
             & \textbf{Global Caption}
             & \textbf{mIoU(\%)} $\uparrow$
             & \textbf{O-mIoU(\%)} $\uparrow$
             & \textbf{SR$_E$(\%)} $\uparrow$
             & \textbf{SR$_R$(\%)} $\uparrow$
             & \textbf{CLIP$_G$} $\uparrow$
             & \textbf{CLIP$_L$} $\uparrow$                   \\
            \midrule

            \multirow{4}{*}{EliGen-SD3}
             & Simple                         & Short & Long
             & \textbf{70.39}
             & \textbf{47.09}
             & \underline{91.83}
             & \textbf{91.06}
             & 36.27
             & 29.15                              \\

             & Simple                         & Short & Short
             & \underline{69.17}
             & \underline{45.80}
             & \textbf{92.95}
             & \underline{90.62}
             & 36.37
             & \textbf{29.66}                                 \\

             & Simple                         & Long  & Long
             & 60.90
             & 35.85
             & 89.55
             & 89.47
             & \underline{36.69}
             & \underline{29.24}                                          \\

             & Simple                         & Long  & Short
             & 56.28
             & 32.32
             & 89.10
             & 89.64
             & \textbf{36.79}
             & 28.62                                          \\
            \midrule

            \multirow{4}{*}{EliGen-SD3}
             & Regular                        & Short & Long
             & \textbf{62.40}
             & \textbf{36.66}
             & \underline{89.45}
             & \underline{86.75}
             & 35.64
             & \underline{28.17}                              \\

             & Regular                        & Short & Short
             & \underline{62.32}
             & \underline{36.33}
             & \textbf{90.72}
             & \textbf{86.94}
             & 35.55
             & \textbf{28.55}                                 \\

             & Regular                        & Long  & Long
             & 53.30
             & 26.64
             & 85.94
             & 84.66
             & \underline{35.99}
             & 28.11                                          \\

             & Regular                        & Long  & Short
             & 48.60
             & 23.48
             & 87.25
             & 85.45
             & \textbf{36.11}
             & 27.47                                          \\
            \midrule

            \multirow{4}{*}{EliGen-SD3}
             & Complex                        & Short & Long
             & \textbf{57.90}
             & \textbf{30.99}
             & \underline{83.30}
             & \underline{85.59}
             & 35.46
             & 26.97                             \\

             & Complex                        & Short & Short
             & \underline{57.48}
             & \underline{30.51}
             & \textbf{85.10}
             & \textbf{85.76}
             & 35.31
             & \textbf{27.54}                                 \\

             & Complex                        & Long  & Long
             & 47.64
             & 21.31
             & 78.49
             & 82.19
             & \textbf{35.85}
             & \underline{27.09}                                        \\

             & Complex                        & Long  & Short
             & 44.23
             & 19.52
             & 79.67
             & 84.25
             & \underline{35.79}
             & 26.43                                          \\

            \bottomrule
        \end{tabular}
    \end{adjustbox}
\end{table*}

\section{Conclusion, Limitations, and Broader Impact}
\label{sec:conclusion_limitations_broader_impact}

We introduced \algoname{}, a large-scale dataset for dense and overlap-aware Layout-to-Image generation. \algoname{} contains approximately 500K images with an average of 6.6 objects per image, exceeding existing datasets by $1.67\times$ in annotation density, and provides caption-rich supervision with short and long captions at both the global and per-object levels. Across multiple benchmarks, state-of-the-art L2I methods trained on \algoname{} show consistent improvements and faster convergence, highlighting the importance of dense and overlap-aware supervision for controllable image generation.

Despite these improvements, \algoname{} has several limitations. Since our pipeline uses MLLMs for captioning and grounding visual entities, the dataset may inherit model biases and annotation noise. In addition, although the \algoname{} dataset focuses on dense scenes with overlapping objects, it may still underrepresent difficult cases such as unusual object interactions, transparent objects, and rare object categories. \algoname{} is intended to support the development of more reliable Layout-to-Image models for real-world and industrial applications, where fine-grained visual control and spatially grounded image synthesis are basic requirements. However, because the dataset is constructed using open-source foundation models, it may inherit the same biases, failure modes, and safety risks present in existing open-source generative models. Further research on the misuse of L2I models is needed to better understand these risks and support the safe deployment of controllable image generation systems.

\paragraph{Acknowledgment} This work is supported by U.S. National Science Foundation Award IIS-2433768 and IIS-2127544. This work also used Delta and DeltaAI at National Center for Supercomputing Applications (NCSA) through allocation CIS260179 from the Advanced Cyberinfrastructure Coordination Ecosystem: Services \& Support (ACCESS) program, which is supported by U.S. National Science Foundation grants \#2138259, \#2138286, \#2138307, \#2137603, and \#2138296. We also thank Lambda, Inc. for their valuable compute resource help.

\bibliographystyle{unsrtnat}
\bibliography{main}

\newpage
\appendix
\section{Technical appendices and supplementary material}
\subsection{Prompts}

To construct OverLay++, we employ Qwen3-VL-32B~\citep{bai2025qwen3vl} throughout the dataset curation pipeline for object detection, scene filtering, and hierarchical caption generation.

The prompts are designed to encourage dense and overlap-aware annotations while seeking to reduce hallucinated object annotations and provide rich semantic descriptions. Specifically, the object detection prompt explicitly encourages overlapping bounding boxes and detailed object-centric descriptions, while the scene filtering stage performs per-object validation to remove hallucinated detections. In addition, we generate both global and local captions at long and short granularities to support different Layout-to-Image architectures with varying text encoder capacities. The exact prompts used in each stage are provided below.

\begin{tcolorbox}[promptbox, title=Object Detection]
    Analyze this image and detect ALL objects that are present in the image. For each detected object, provide:\\
    1. The object name/category\\
    2. A bounding box [x1, y1, x2, y2] indicating where the object is located in the image\\
    3. A detailed local\_prompt description that describes the object's appearance, attributes, style, and visual characteristics\\[4pt]
    IMPORTANT REQUIREMENTS:\\
    \textbullet\ Detect objects that are ACTUALLY present in the image\\
    \textbullet\ CRITICAL: Each object's bounding box MUST have some overlap area with at least one other object's bounding box\\
    \textbullet\ Bounding boxes must NOT be very small --- each bounding box should have a minimum width and height of at least $5\%$ of the image resolution\\
    \textbullet\ Minimum bounding box dimensions: width $\geq 51$ pixels, height $\geq 51$ pixels\\
    \textbullet\ Overlapping means the bounding boxes should share at least some common area (intersection over union $> 0$ or at least partial overlap)\\
    \textbullet\ Include various types of objects: foreground objects, background objects, parts of larger objects, etc.\\
    \textbullet\ The local\_prompt should be a detailed text description that includes:\\
    \hspace*{1.5em}* Visual appearance\\
    \hspace*{1.5em}* Size and proportions\\
    \hspace*{1.5em}* Style and design details\\
    \hspace*{1.5em}* Any specific attributes or features\\
    \hspace*{1.5em}* Contextual details relevant to the scene\\[4pt]
    Response in JSON format:\\
    \texttt{\{}\\
    \hspace*{1em}\texttt{"detections": [}\\
    \hspace*{2em}\texttt{\{}\\
    \hspace*{3em}\texttt{"category": "object\_category",}\\
    \hspace*{3em}\texttt{"bbox": [x1, y1, x2, y2],}\\
    \hspace*{3em}\texttt{"local\_prompt": "detailed description of the object's appearance, style, and visual characteristics"}\\
    \hspace*{2em}\texttt{\},}\\
    \hspace*{2em}\texttt{...}\\
    \hspace*{1em}\texttt{]}\\
    \texttt{\}}\\[4pt]
    The bounding box coordinates should be in the range $[0, 1024]$ where $(0,0)$ is the top-left corner.\\
    The image dimensions are: width$=1024$, height$=1024$.\\
    CRITICAL: Ensure all bounding boxes have overlapping areas with at least one other bounding box.\\
    CRITICAL: Do not create very small bounding boxes.\\
    Strictly follow the JSON format without any additional text.
\end{tcolorbox}

\begin{tcolorbox}[promptbox, title=Scene Filtering]
    Look at this cropped image region. Is the following object present in this image?\\[4pt]
    \textbf{Object:} \{category\}. \{local\_prompt\}.\\[4pt]
    Answer with only `yes' or `no'. No additional explanation.
\end{tcolorbox}

\begin{tcolorbox}[promptbox, title=Global Caption]
    Describe this image in one or two sentences, including the main objects, activities, and setting.
\end{tcolorbox}

\begin{tcolorbox}[promptbox, title=Short Global Caption]
    Describe this image with a short global caption.\\[4pt]
    \textbf{Requirements:}\\
    \textbullet\ Summarize the overall scene.\\
    \textbullet\ Include the key object(s) and setting.\\
    \textbullet\ Keep it under 20 words.\\[4pt]
    Return only the caption text with no additional explanation.
\end{tcolorbox}

\begin{tcolorbox}[promptbox, title=Short Local Prompt]
    You are given a full image and object detections.\\
    Use the image content to answer (not detections text alone).\\
    For each detection, return:\\
    1) short\_local\_prompt: less than 20 words\\[4pt]
    \textbf{IMPORTANT:}\\
    \textbullet\ Keep the same order and number of detections.\\
    \textbullet\ Each short\_local\_prompt must include color and any other visual attributes.\\
    \textbullet\ Each short\_local\_prompt must be less than 20 words.\\
    \textbullet\ Return only JSON in this format:\\
    \texttt{\{}\\
    \hspace*{1em}\texttt{"detections": [}\\
    \hspace*{2em}\texttt{\{"short\_local\_prompt": "..."\},}\\
    \hspace*{2em}\texttt{...}\\
    \hspace*{1em}\texttt{]}\\
    \texttt{\}}\\[4pt]
    Detections input: \texttt{<JSON-serialized list of detections>}
\end{tcolorbox}

\subsection{Long Local Captions with a Long-Context Text Encoder}
\label{sec:supp_flux_ablation}

The degraded performance with long local prompts in Table~\ref{tab:ablation_prompt_granularity} may reflect truncation in EliGen-SD3's local-prompt encoder path. The CLIP path has a 77-token limit~\citep{radford2021learningtransferablevisualmodels}, while long local captions average 89.70 words, making truncation plausible. This is a hypothesis rather than a direct measurement of truncation's effect.
\begin{table*}[t]
    \centering
    \small
    \setlength{\tabcolsep}{4pt}
    \renewcommand{\arraystretch}{1.22}
    \caption{
        \textbf{FLUX results on OverLayBench.}
        We fine-tune the released EliGen-FLUX model for 5K steps (batch size 8) on OverLay++ with long local captions. Our model improves over the baseline on nearly every metric across all complexity settings.
        Improvements over the baseline are shown in {\color{ForestGreen}green}, and \textbf{bold} indicates the best result in each column within each split.
    }
    \label{tab:flux_overlaybench}
    \begin{adjustbox}{max width=\textwidth}
        \begin{tabular}{@{}lccccccc@{}}
            \toprule
            \textbf{Method}
             & \textbf{mIoU(\%)} $\uparrow$
             & \textbf{O-mIoU(\%)} $\uparrow$
             & \textbf{SR$_E$(\%)} $\uparrow$
             & \textbf{SR$_R$(\%)} $\uparrow$
             & \textbf{CLIP$_G$} $\uparrow$
             & \textbf{CLIP$_L$} $\uparrow$
             & \textbf{FID} $\downarrow$ \\
            \midrule
 
            \multicolumn{8}{c}{\textbf{OverLayBench-Simple}} \\
            \midrule
            EliGen-FLUX~\citep{zhang2025eligenentitylevelcontrolledimage}
             & 68.17 & 43.72 & 86.50 & 89.67 & 36.65 & 28.29 & 28.87 \\
 
            EliGen-FLUX (Ours)
             & \textbf{69.32} {\color{ForestGreen}(+1.69\%)}
             & \textbf{47.21} {\color{ForestGreen}(+7.98\%)}
             & \textbf{91.32} {\color{ForestGreen}(+5.57\%)}
             & \textbf{91.31} {\color{ForestGreen}(+1.83\%)}
             & \textbf{36.81} {\color{ForestGreen}(+0.44\%)}
             & \textbf{29.77} {\color{ForestGreen}(+5.23\%)}
             & \textbf{26.42} {\color{ForestGreen}(-8.49\%)} \\
 
            \midrule\midrule
 
            \multicolumn{8}{c}{\textbf{OverLayBench-Regular}} \\
            \midrule
            EliGen-FLUX~\citep{zhang2025eligenentitylevelcontrolledimage}
             & 58.56 & 32.62 & 80.85 & 84.42 & \textbf{36.27} & 27.05 & 45.65 \\
 
            EliGen-FLUX (Ours)
             & \textbf{60.36} {\color{ForestGreen}(+3.07\%)}
             & \textbf{35.72} {\color{ForestGreen}(+9.50\%)}
             & \textbf{88.64} {\color{ForestGreen}(+9.64\%)}
             & \textbf{87.13} {\color{ForestGreen}(+3.21\%)}
             & 36.20
             & \textbf{28.70} {\color{ForestGreen}(+6.10\%)}
             & \textbf{43.51} {\color{ForestGreen}(-4.69\%)} \\
 
            \midrule\midrule
 
            \multicolumn{8}{c}{\textbf{OverLayBench-Complex}} \\
            \midrule
            EliGen-FLUX~\citep{zhang2025eligenentitylevelcontrolledimage}
             & 52.53 & 26.19 & 74.03 & 84.09 & 36.18 & 25.92 & 50.41 \\
 
            EliGen-FLUX (Ours)
             & \textbf{54.74} {\color{ForestGreen}(+4.21\%)}
             & \textbf{29.03} {\color{ForestGreen}(+10.84\%)}
             & \textbf{82.57} {\color{ForestGreen}(+11.54\%)}
             & \textbf{86.81} {\color{ForestGreen}(+3.23\%)}
             & \textbf{36.21} {\color{ForestGreen}(+0.08\%)}
             & \textbf{27.84} {\color{ForestGreen}(+7.41\%)}
             & \textbf{47.53} {\color{ForestGreen}(-5.71\%)} \\
 
            \bottomrule
        \end{tabular}
    \end{adjustbox}
\end{table*}
\begin{table}[t]
    \centering
    \small
    \setlength{\tabcolsep}{3pt}
    \renewcommand{\arraystretch}{1.25}
    \caption{
        \textbf{FLUX results on LayoutSAM-Eval.}
        Comparison of region-wise and global quality for FLUX-based methods. Our model improves over EliGen-FLUX on all region-wise metrics while also improving global quality.
        Improvements over EliGen-FLUX are shown in {\color{ForestGreen}green}, and \textbf{bold} and \underline{underline} denote the best and second-best results among all compared methods.
    }
    \label{tab:flux_layoutsam}
    \begin{adjustbox}{max width=\linewidth}
        \begin{tabular}{lcccccc}
            \toprule
             & \multicolumn{4}{c}{\textbf{Region-wise Quality}}
             & \multicolumn{2}{c}{\textbf{Global Quality}} \\
            \cmidrule(lr){2-5} \cmidrule(lr){6-7}
            \textbf{Model}
             & \textbf{Spatial(\%)} $\uparrow$
             & \textbf{Color(\%)} $\uparrow$
             & \textbf{Texture(\%)} $\uparrow$
             & \textbf{Shape(\%)} $\uparrow$
             & \textbf{FID} $\downarrow$
             & \textbf{IS} $\uparrow$ \\
            \midrule
            FLUX~\citep{BlackForestLabs2024Flux}
             & 79.61 & 62.54 & 64.41 & 62.90 & \underline{20.78} & \textbf{22.15} \\
            \midrule
            EliGen-FLUX~\citep{zhang2025eligenentitylevelcontrolledimage}
             & \underline{93.25} & \underline{79.95} & \underline{82.96} & \underline{82.99} & 28.99 & 18.23 \\
 
            EliGen-FLUX (Ours)
             & \textbf{96.18} {\color{ForestGreen}(+3.14\%)}
             & \textbf{90.81} {\color{ForestGreen}(+13.58\%)}
             & \textbf{93.23} {\color{ForestGreen}(+12.38\%)}
             & \textbf{92.79} {\color{ForestGreen}(+11.81\%)}
             & \textbf{15.98} {\color{ForestGreen}(-44.88\%)}
             & \underline{20.58} {\color{ForestGreen}(+12.89\%)} \\
            \bottomrule
        \end{tabular}
    \end{adjustbox}
\end{table}
 
To test long local captions directly, we fine-tune the released EliGen-FLUX model, whose local prompts use a longer-context T5 path, on the OverLay++ dataset with long local captions. Due to limited compute, we train for 5K steps with a batch size of 8. The FLUX results are consistent with the possibility that a longer-context local-prompt path can use more of the caption detail. As shown in Table~\ref{tab:flux_overlaybench}, our model improves over the EliGen-FLUX baseline on nearly every metric across all three OverLayBench splits, with large gains on region-aware metrics (e.g., SR$_E$ improves by up to 11.54\% on the Complex split) as well as better FID. The only exception is a marginal drop in CLIP$_G$ on the Regular split. Table~\ref{tab:flux_layoutsam} shows the same trend on LayoutSAM-Eval: our model improves over EliGen-FLUX on all region-wise metrics (Spatial, Color, Texture, Shape) as well as on FID and IS.

\subsection{Global Alignment with Long-CLIP}
\label{sec:supp_longclip}
\begin{table}[t]
    \centering
    \caption{\textbf{Global text--image alignment with Long-CLIP on OverLayBench.} Higher scores indicate stronger alignment. These scores use a different encoder from CLIP$_G$ in the main tables and are not directly comparable in magnitude.}
    \label{tab:longclip_overlaybench}
    \begin{tabular}{llc}
        \toprule
        \textbf{Split} & \textbf{Model} & \textbf{Long-CLIP$_G$} $\uparrow$ \\
        \midrule
        Simple  & EliGen-SD3        & 0.2564 \\
                & EliGen-SD3 (Ours) & \textbf{0.2573} \\
        Regular & EliGen-SD3        & 0.2556 \\
                & EliGen-SD3 (Ours) & \textbf{0.2559} \\
        Complex & EliGen-SD3        & 0.2563 \\
                & EliGen-SD3 (Ours) & \textbf{0.2570} \\
        \bottomrule
    \end{tabular}
\end{table}

The global prompts in OverLayBench average 79.66, 86.81, and 82.91 words on the Simple, Regular, and Complex splits, respectively, exceeding the 77-token input limit of CLIP. We therefore reevaluate global text--image alignment with Long-CLIP~\citep{zhang2024longclip}, which supports up to 248 tokens. As shown in Table~\ref{tab:longclip_overlaybench}, the OverLay++-trained EliGen-SD3 model has slightly higher Long-CLIP$_G$ scores than its baseline on all three splits. This comparison changes the evaluation encoder; it does not test a change to the generation model's local-prompt encoder.

\subsection{Source of the Improvements}
\label{sec:supp_source_improvements}
\begin{table*}[t]
    \centering
    \small
    \setlength{\tabcolsep}{4pt}
    \renewcommand{\arraystretch}{1.22}
    \caption{
        \textbf{Ablation of object count and local caption length on OverLayBench.}
        All models use the SD3 backbone and are trained for 2K steps on \algoname{} subsets with matched mean IoU. ``Obj.'' and ``Cap.'' denote the mean number of objects per layout and the mean local caption length (in words) of the training data. Improvements over the EliGen-SD3 baseline are shown in {\color{ForestGreen}green}, and \textbf{bold} and \underline{underline} indicate the best and second-best results in each column.
    }
    \label{tab:supp_ablation_source}
    \begin{adjustbox}{max width=\textwidth}
        \begin{tabular}{@{}lccccccccc@{}}
            \toprule
            \textbf{Method}
             & \textbf{Obj.}
             & \textbf{Cap.}
             & \textbf{mIoU(\%)} $\uparrow$
             & \textbf{O-mIoU(\%)} $\uparrow$
             & \textbf{SR$_E$(\%)} $\uparrow$
             & \textbf{SR$_R$(\%)} $\uparrow$
             & \textbf{CLIP$_G$} $\uparrow$
             & \textbf{CLIP$_L$} $\uparrow$
             & \textbf{FID} $\downarrow$ \\
            \midrule
 
            \multicolumn{10}{c}{\textbf{OverLayBench-Simple}} \\
            \midrule
            EliGen-SD3~\citep{zhang2025eligenentitylevelcontrolledimage}
             & 2.56 & 2.12
             & 58.70 & 32.54 & 80.32 & 86.01
             & \textbf{36.88} & 28.04 & \textbf{31.16} \\
 
            EliGen-SD3 (Ours, incr.\ objects)
             & 6.68 & 1.88
             & \underline{66.30} {\color{ForestGreen}(+12.95\%)}
             & \underline{39.58} {\color{ForestGreen}(+21.63\%)}
             & \underline{90.36} {\color{ForestGreen}(+12.50\%)}
             & \underline{89.98} {\color{ForestGreen}(+4.62\%)}
             & \underline{36.51}
             & \textbf{29.66} {\color{ForestGreen}(+5.78\%)}
             & \underline{32.05} \\
 
            EliGen-SD3 (Ours, incr.\ caption len)
             & 6.59 & 15.89
             & \textbf{68.72} {\color{ForestGreen}(+17.07\%)}
             & \textbf{43.97} {\color{ForestGreen}(+35.13\%)}
             & \textbf{91.62} {\color{ForestGreen}(+14.07\%)}
             & \textbf{90.28} {\color{ForestGreen}(+4.96\%)}
             & 36.19
             & \underline{29.21} {\color{ForestGreen}(+4.17\%)}
             & 36.31 \\
 
            \midrule\midrule
 
            \multicolumn{10}{c}{\textbf{OverLayBench-Regular}} \\
            \midrule
            EliGen-SD3~\citep{zhang2025eligenentitylevelcontrolledimage}
             & 2.56 & 2.12
             & 49.04 & 22.38 & 76.19 & 81.41
             & \textbf{36.54} & 26.55 & \textbf{49.07} \\
 
            EliGen-SD3 (Ours, incr.\ objects)
             & 6.68 & 1.88
             & \underline{58.74} {\color{ForestGreen}(+19.78\%)}
             & \underline{31.56} {\color{ForestGreen}(+41.02\%)}
             & \underline{88.09} {\color{ForestGreen}(+15.62\%)}
             & \underline{85.74} {\color{ForestGreen}(+5.32\%)}
             & \underline{35.88}
             & \textbf{28.56} {\color{ForestGreen}(+7.57\%)}
             & \underline{50.99} \\
 
            EliGen-SD3 (Ours, incr.\ caption len)
             & 6.59 & 15.89
             & \textbf{60.72} {\color{ForestGreen}(+23.82\%)}
             & \textbf{33.53} {\color{ForestGreen}(+49.82\%)}
             & \textbf{89.66} {\color{ForestGreen}(+17.68\%)}
             & \textbf{85.93} {\color{ForestGreen}(+5.55\%)}
             & 35.35
             & \underline{28.18} {\color{ForestGreen}(+6.14\%)}
             & 59.65 \\
 
            \midrule\midrule
 
            \multicolumn{10}{c}{\textbf{OverLayBench-Complex}} \\
            \midrule
            EliGen-SD3~\citep{zhang2025eligenentitylevelcontrolledimage}
             & 2.56 & 2.12
             & 42.87 & 16.87 & 67.22 & 79.75
             & \textbf{36.44} & 25.62 & \textbf{53.52} \\
 
            EliGen-SD3 (Ours, incr.\ objects)
             & 6.68 & 1.88
             & \underline{54.15} {\color{ForestGreen}(+26.31\%)}
             & \underline{26.06} {\color{ForestGreen}(+54.48\%)}
             & \underline{82.33} {\color{ForestGreen}(+22.48\%)}
             & \textbf{84.83} {\color{ForestGreen}(+6.37\%)}
             & \underline{35.66}
             & \textbf{27.63} {\color{ForestGreen}(+7.85\%)}
             & \underline{55.65} \\
 
            EliGen-SD3 (Ours, incr.\ caption len)
             & 6.59 & 15.89
             & \textbf{56.00} {\color{ForestGreen}(+30.63\%)}
             & \textbf{27.91} {\color{ForestGreen}(+65.44\%)}
             & \textbf{83.66} {\color{ForestGreen}(+24.46\%)}
             & \underline{84.27} {\color{ForestGreen}(+5.67\%)}
             & 35.14
             & \underline{27.27} {\color{ForestGreen}(+6.44\%)}
             & 64.30 \\
 
            \bottomrule
        \end{tabular}
    \end{adjustbox}
\end{table*}

We examine the effects of object count and local caption length using two ablations on the SD3 backbone, each trained for 2K steps. For the 450K-image ablation subset, we remove the 50K OverLay++ images with the highest nearest-neighbor IoU, reducing its mean from 0.31 in the full dataset to 0.27, close to EliGen's 0.26. Dataset composition and object count are not identical across all settings, so these comparisons do not isolate either factor perfectly. Results on the three OverLayBench splits (Simple, Regular, Complex) are reported in Table~\ref{tab:supp_ablation_source}.

We compare EliGen-SD3, trained on the EliGen dataset (2.56 objects/layout, 2.12-word local captions), against EliGen-SD3 (Ours, incr.\ objects), trained on the OverLay++ dataset with local captions restricted to a brief object description (mean 1.88 words, \emph{shorter} than EliGen's). The settings differ in object count (6.68 vs.\ 2.56 objects/layout), caption length, and potentially other dataset properties; Table~\ref{tab:dataset_analysis} reports different nearest-neighbor IoU values for the full datasets. Despite using shorter captions, training on OverLay++ consistently outperforms EliGen on all local instruction-following metrics (mIoU, O-mIoU, SR$_E$, SR$_R$, and CLIP$_L$) across all three OverLayBench splits.

We then train EliGen-SD3 (Ours, incr.\ caption len), which replaces the brief object descriptions above (mean 1.88 words) with the short local captions of the OverLay++ dataset (mean 15.89 words), with a slightly different mean object count (6.59 rather than 6.68). The longer-caption setting improves mIoU, O-mIoU, and SR$_E$ across all three splits. SR$_R$ improves on Simple and Regular but decreases on Complex; CLIP$_L$ decreases across all three. Both settings use 2K training steps, so these differences cannot be attributed to fewer steps.

CLIP$_G$ decreases and FID worsens in the longer-caption 2K-step setting; for example, FID rises from 50.99 to 59.65 on Regular. These results do not establish the cause of the global-metric changes. The longer training runs in Table~\ref{tab:layoutsam_eval} and Table~\ref{tab:quantitative_results_all_training}, and the FLUX ablation in Section~\ref{sec:supp_flux_ablation}, report better global metrics, but these comparisons do not isolate which change is responsible.

\begin{table}[t]
    \centering
    \small
    \caption{\textbf{LayoutSAM-Eval during EliGen-SD3 training.} The pretrained SD3 backbone is shown for context. These 5K- and 10K-step runs use the main training settings and are separate from the 2K-step, overlap-matched ablation in Table~\ref{tab:supp_ablation_source}.}
    \label{tab:supp_layoutsam_checkpoints}
    \begin{tabular}{lccccccc}
        \toprule
        \textbf{Model} & \textbf{Steps} & \textbf{Spatial} $\uparrow$ & \textbf{Color} $\uparrow$ & \textbf{Texture} $\uparrow$ & \textbf{Shape} $\uparrow$ & \textbf{FID} $\downarrow$ & \textbf{IS} $\uparrow$ \\
        \midrule
        SD3 (pretrained) & --  & 78.78 & 65.94 & 68.43 & 67.07 & 20.68 & 21.74 \\
        EliGen-SD3       & 5K  & 93.14 & 82.13 & 86.60 & 86.15 & 25.88 & 19.20 \\
        EliGen-SD3 (Ours)& 5K  & \textbf{95.61} & \textbf{90.86} & \textbf{93.12} & \textbf{92.94} & \textbf{19.54} & \textbf{19.94} \\
        EliGen-SD3       & 10K & 93.04 & 82.96 & 86.73 & 86.35 & 25.17 & 19.21 \\
        EliGen-SD3 (Ours)& 10K & \textbf{95.45} & \textbf{90.03} & \textbf{93.14} & \textbf{92.81} & \textbf{19.61} & \textbf{19.54} \\
        \bottomrule
    \end{tabular}
\end{table}

For completeness, Table~\ref{tab:supp_layoutsam_checkpoints} reports LayoutSAM-Eval scores at 5K and 10K steps for the main EliGen-SD3 training runs. OverLay++ improves the region-wise metrics and FID relative to EliGen at both checkpoints, while IS is comparable. These runs use the main training data and captions, so they do not extend the overlap-matched 2K-step ablation or identify the individual contribution of object count, overlap, and caption detail.

\subsection{Human Evaluation of Local Layout and Instruction Following}
\label{sec:supp_human_study}
\begin{table}[t]
    \centering
    \small
    \setlength{\tabcolsep}{6pt}
    \renewcommand{\arraystretch}{1.25}
    \caption{
        \textbf{Human preference study on local layout and instruction following.}
        Percentage of bounding boxes whose generated region was judged to satisfy both the box and its local prompt (``yes'' rate). Improvements over the baseline are shown in {\color{ForestGreen}green}, and \textbf{bold} indicates the best result in each row.
    }
    \label{tab:human_study}
    \begin{adjustbox}{max width=\linewidth}
        \begin{tabular}{lccc}
            \toprule
            \textbf{Split}
             & \textbf{EliGen-SD3} $\uparrow$
             & \textbf{EliGen-SD3 (Ours)} $\uparrow$
             & \textbf{Difference (pp)} \\
            \midrule
            All
             & 69.90 & \textbf{81.70} & {\color{ForestGreen}+11.80} \\
            \midrule
            Simple
             & 73.70 & \textbf{83.00} & {\color{ForestGreen}+9.30} \\
            Regular
             & 69.70 & \textbf{82.80} & {\color{ForestGreen}+13.10} \\
            Complex
             & 68.20 & \textbf{80.00} & {\color{ForestGreen}+11.80} \\
            \bottomrule
        \end{tabular}
    \end{adjustbox}
\end{table}
 
To directly evaluate local layout and instruction-following quality, we conduct a human yes/no evaluation using the EliGen method. We compare images generated by models trained on the EliGen dataset (baseline) and on our proposed OverLay++ dataset.
 
We sample 60 layouts from the OverLayBench Simple, Regular, and Complex splits, spanning 780 bounding boxes in total. The boxes are rated by 12 raters, with an average of 2.37 ratings per box. For each box, raters judge whether the generated region satisfies the constraint imposed by its bounding box and corresponding local prompt, answering yes or no. We report the fraction of ``yes'' ratings for each model, overall and per split, in Table~\ref{tab:human_study}.
 
The model trained on OverLay++ receives a higher fraction of ``yes'' ratings than the baseline, both overall (+11.80 percentage points) and within every difficulty split, with gains of +9.30, +13.10, and +11.80 points on Simple, Regular, and Complex, respectively. This provides direct human-judgment evidence, complementing the automatic benchmarks in the main text, that training on OverLay++ substantially improves local layout and instruction-following capability.

\subsection{Dataset Composition}
\label{sec:supp_dataset_composition}
\begin{table*}[t]
    \centering
    \small
    \setlength{\tabcolsep}{5pt}
    \renewcommand{\arraystretch}{1.2}
    \caption{
        \textbf{Composition of OverLay++ by semantic group.}
        Short local prompts are zero-shot classified into 26 broad semantic groups. Count is the number of bounding boxes in each group, and \% is its share of all bounding boxes.
    }
    \label{tab:dataset_composition}
    \begin{adjustbox}{max width=\textwidth}
        \begin{tabular}{@{}lrr@{\hspace{2em}}lrr@{}}
            \toprule
            \textbf{Group} & \textbf{Count} & \textbf{\%}
             & \textbf{Group} & \textbf{Count} & \textbf{\%} \\
            \midrule
            Furniture and home decor   & 499,498 & 15.2 & Fences and stairs      & 71,097 & 2.2 \\
            Hair and hairstyles        & 346,341 & 10.5 & Weapons and armor      & 66,173 & 2.0 \\
            Clothing and accessories   & 316,170 &  9.6 & Signs and text         & 64,401 & 2.0 \\
            Sky and clouds             & 287,859 &  8.7 & Food and drinks        & 59,890 & 1.8 \\
            Flowers and gardens        & 209,013 &  6.3 & Musical instruments    & 58,097 & 1.8 \\
            Walls and floors           & 169,690 &  5.2 & Snow and ice           & 49,853 & 1.5 \\
            Trees and plants           & 166,339 &  5.1 & Electronics and screens & 47,520 & 1.4 \\
            Buildings and architecture & 157,008 &  4.8 & Roads and streets      & 31,766 & 1.0 \\
            Mountains and landscape    & 147,946 &  4.5 & Birds and insects      & 31,573 & 1.0 \\
            People and faces           & 142,364 &  4.3 & Animals and pets       & 24,086 & 0.7 \\
            Water and ocean            & 105,980 &  3.2 & Fire and smoke         & 21,836 & 0.7 \\
            Sports equipment           &  97,876 &  3.0 & Books and papers       & 20,771 & 0.6 \\
            Art and sculptures         &  78,429 &  2.4 & Vehicles and transport & 20,414 & 0.6 \\
            \bottomrule
        \end{tabular}
    \end{adjustbox}
\end{table*}
 
Existing Layout-to-Image datasets are largely adapted from object detection or grounding datasets, and therefore come with explicit class labels drawn from a fixed set of categories. In contrast, OverLay++ is open-domain: object categories are generated directly by Qwen3-VL-32B as free-form text during detection, rather than selected from a predefined class list.
 
To give a sense of the dataset's composition, we perform zero-shot classification of the short local prompts into 26 broad semantic groups. Table~\ref{tab:dataset_composition} reports the number of bounding boxes in each group, where \emph{count} refers to the number of bounding boxes belonging to that group, along with its share of all boxes. The dataset covers a broad range of content, from furniture and home decor, clothing and accessories, and hair and hairstyles to surfaces (walls and floors) and natural scenery (sky and clouds, flowers and gardens, mountains and landscape), with a long tail of less frequent categories such as musical instruments, weapons and armor, and vehicles.

\end{document}